\documentclass[a4paper,11pt]{article}
\usepackage{jinstpub} 
\usepackage{lineno}
\usepackage{graphicx}
\usepackage[symbol]{footmisc}
\usepackage{soul}
\newcommand\DLweak{$0.3223\pm0.0765$}
\newcommand\DLmix{$0.5748\pm0.1358$}
\newcommand\DLgood{$0.7710\pm0.0499$}

\title{\boldmath Weakly supervised neural network: segmentation of complex structures in X-ray microCT}

\author[a,1,2]{Daniele Rusconi,\note[1]{These authors equally contributed to this work.}\note[2]{Corresponding author.}}
\author[b,1]{Michela Ascolese}
\author[a]{Stephanie Fest-Santini,}
\author[b,c,d,e]{Alberto Bravin,}
\author[a]{Maurizio Santini}

\affiliation[a]{Department of Engineering and Applied Sciences, University of Bergamo\\Viale Marconi 20, 24044 Dalmine, Italy}
\affiliation[b]{Department of Physics ‘‘G. Occhialini’’, University of Milano-Bicocca\\Piazza della Scienza 3, 20126 Milan, Italy}
\affiliation[c]{INFN Milano-Bicocca\\Piazza della Scienza 3, 20126 Milan, Italy}
\affiliation[d]{Department of Physics and STAR Research Infrastructure, University of Calabria\\Via Tito Flavio, 87036 Rende, CS, Italy}
\affiliation[e]{CNR-NANOTEC, SS di Rende\\Via Pietro Bucci, 87036 Rende, CS, Italy}

\emailAdd{daniele.rusconi@unibg.it}

\abstract{
Segmentation of complex structures in X-ray tomographic data is a fundamental task in biomedical research, but it often requires large amounts of precisely annotated data, making fully supervised approaches costly and difficult to scale. In this study, weakly supervised deep learning is investigated as a strategy to reduce annotation effort while maintaining accurate segmentation.

A two-dimensional convolutional neural network based on the nnU-Net framework was adapted to a weak supervision setting using sparse dot-based annotations, complemented by a limited number of fully segmented images. The approach was evaluated on high-resolution microCT slices of rat kidneys, targeting the segmentation of renal glomeruli, which are small, low-contrast anatomical structures.

Results indicate that weak supervision provides a meaningful learning signal, enabling reliable localization of glomeruli even in the absence of dense labels. Incorporating a small set of high-quality annotations substantially improves segmentation performance, approaching that of a fully supervised model. These findings highlight the potential of weakly supervised learning as an annotation-efficient strategy for the analysis of complex structures in X-ray tomographic data, and suggest that alternative loss formulations tailored to sparse annotations may further enhance performance.
}


\arxivnumber{1234.56789} 

\begin{document}
\maketitle
\flushbottom


\section{Introduction}
\label{sec:intro}

The study of renal glomeruli is central to understanding kidney function and disease progression. These quasi-spherical structures, typically tens of micrometers in diameter, are responsible for blood filtration and fluid regulation. Quantitative and morphological analysis of glomeruli provides important information for fundamental and preclinical research. Histology provides detailed microscopic visualization of tissue structure, enabling the identification of cellular changes, disease diagnosis, and the study of organ function; however, conventional histological techniques are time-consuming, destructive, and prone to sampling bias, motivating the adoption of non-invasive imaging modalities such as X-ray micro-computed tomography (microCT)~\cite{a}.

MicroCT enables high-resolution three-dimensional imaging of intact biological samples, allowing volumetric analysis without physical sectioning. However, segmentation of glomeruli in microCT data remains challenging due to their small size and low contrast relative to surrounding tissue. As a result, classical image processing techniques are often inadequate, while fully supervised deep learning (DL) approaches require dense pixel-level annotations that are costly and time-consuming to obtain~\cite{b}.

Weakly supervised DL offers an alternative by enabling learning from sparse or imprecise annotations. Compared to fully supervised methods, weak supervision relies on inexpensive forms of labeling, such as point annotations or coarse region indicators, thereby reducing annotation effort while retaining semantic information. Previous studies~\cite{c,d} have shown that such approaches can achieve competitive performance applications where dense labeling is impractical.

In this study, an nnU-Net-based framework adapted to weak supervision is employed for the automated segmentation of glomeruli in rat kidney microCT data.

The contribution of this work lies in investigating the trade-off between annotation effort and segmentation performance, demonstrating that competitive results can be achieved using sparse annotations without requiring exhaustive pixel-wise labels.


\section{Data acquisition}
\label{sec:data}

Three male Sprague‑Dawley (SD) rats aged 14 weeks were used in this study. All procedures involving animals were performed in accordance with institutional guidelines in compliance with national (D.L.n.26, March 4, 2014), and international laws and policies (directive 2010/63/EU on the protection of animals used for scientific purposes). This study was approved by the Institutional Animal Care and Use Committees of Istituto di Ricerche Farmacologiche Mario Negri IRCCS and by the Italian Ministry of Health (approval number 991/2023-PR).
Three male SD rats were housed in a specific pathogen-free facility at constant temperature, under a 12:12-h light/dark cycle, with free access to food and water. At 14 weeks of age, rats were anesthetized with isoflurane and maintained under stable anesthesia. The left kidney was then perfused with a radio-opaque silicone polymer (Microfil MV-122; Flow Tech, Carver, MA, USA) according to a previously described protocol~\cite{k}. To minimize beam hardening effects, as indicated by preliminary tests, excised kidneys were preserved in 10\% formalin within centrifuge tubes throughout handling and scanning.

\begin{figure}[h]
\centering
\includegraphics[width=0.4\textwidth]{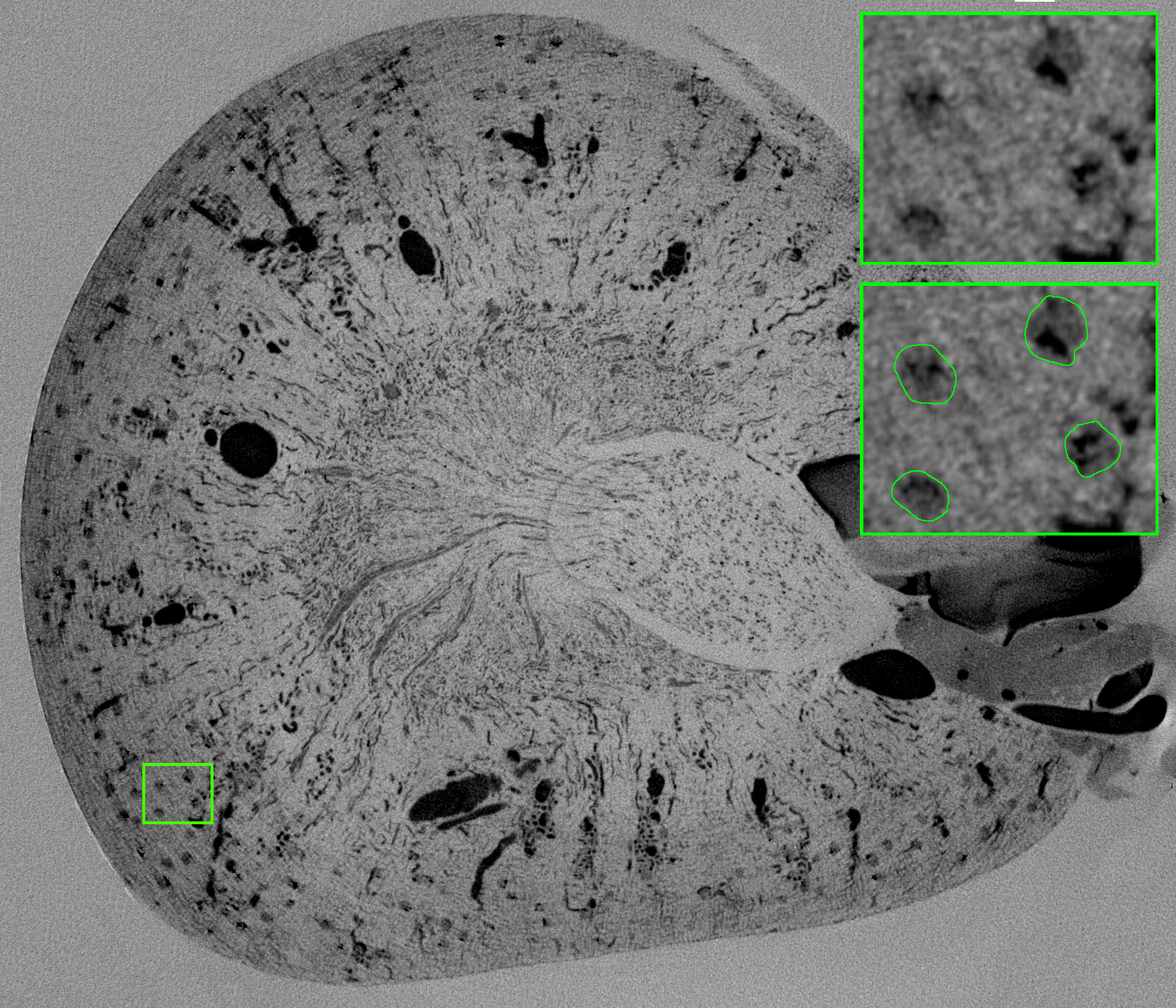}
\includegraphics[width=0.4\textwidth]{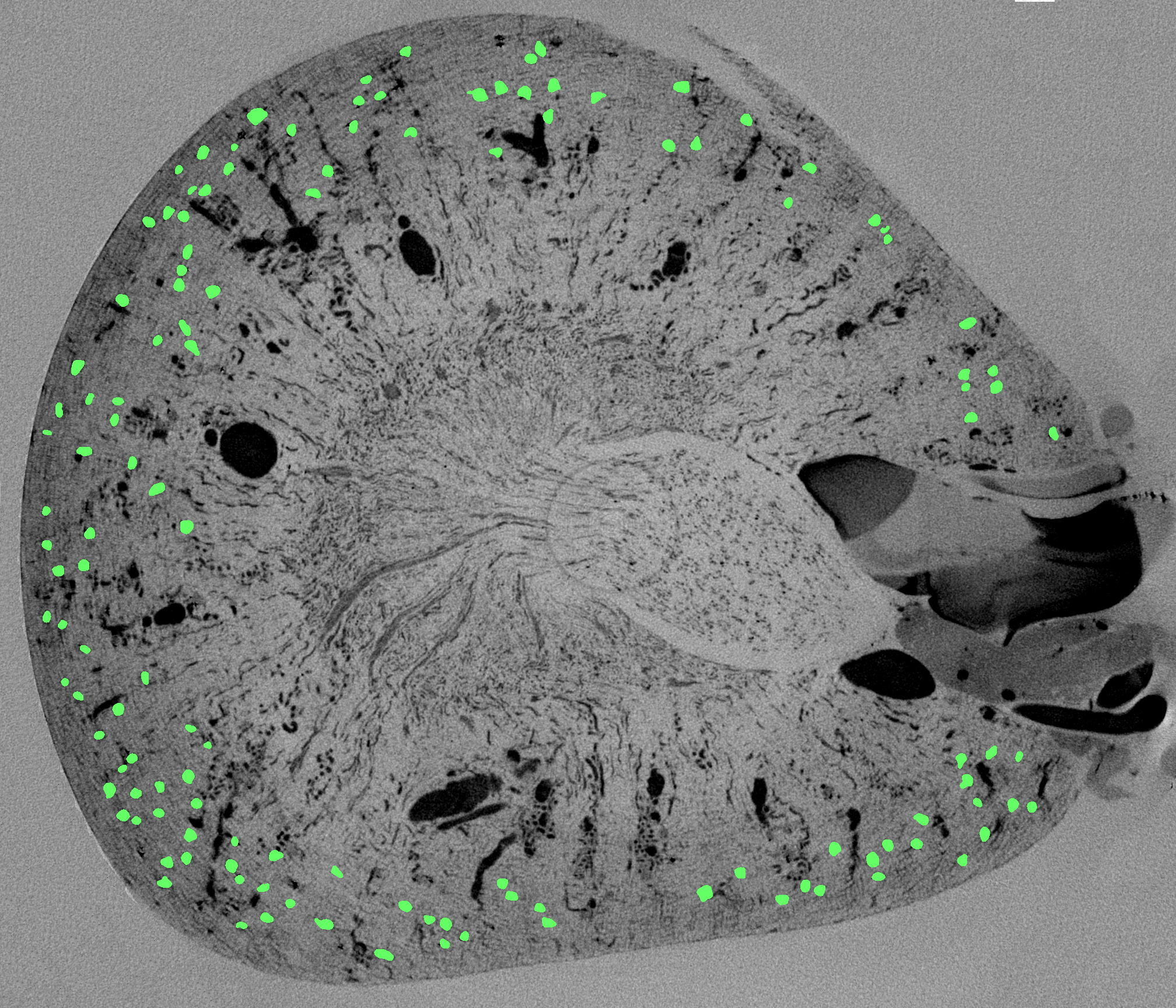}
\caption{Tomographic reconstruction of a rat kidney sample, shown as a single axial slice (left) and a visualization of the spatial distribution of renal glomeruli (right). Detailed views within the left panel illustrate glomeruli at higher resolution (top) and their corresponding positional annotations (bottom).}
\label{fig:sample57}
\end{figure}

MicroCT imaging was performed at an isotropic resolution of $6.81\times6.81\times6.81~\mu$m$^3$ using an open‑type X‑ray source (X-Ray WorX XWT-190-TCNF) operating at 120\,kV and 20\,$\mu$A in high‑power mode. A 16‑bit CMOS detector (PerkinElmer XRD 1611) was used with an acquisition time of 1.8\,s per frame, and 3200 projections were acquired per tomography. During acquisition, the raw projections were corrected for detector charge accumulation (dark‑field correction) and normalized by bright‑field correction, which compensates for spatial non‑uniformity in detector sensitivity and X‑ray source flux density. The normalized projections were then reconstructed using the Feldkamp–Davis–Kress (FDK) algorithm implemented in VGSTUDIO MAX to obtain the three‑dimensional volumes. The specified isotropic resolution was determined via metrological calibration following the procedures described in Santini et al.~\cite{e}.

Images were processed to reduce artifacts and standardize intensity distributions. For each slice, intensity normalization was applied to mitigate residual effects of the contrast fluid and to ensure consistent pixel value ranges across the dataset.


\section{Methods}
\label{sec:methods}

\subsection{Dataset preparation}
\label{sec:dataset_annotations}

Three samples were analyzed, resulting in approximately 6000 images. A subset of these images was annotated and used to train and evaluate the DL model under different supervision regimes.

The fully segmented dataset consisted of approximately 300 slices that were manually segmented by a trained professional using the Dragonfly software platform. In this workflow, AI-assisted tools provided preliminary segmentations, which were subsequently refined by the expert, resulting in accurate pixel-wise masks. On average, between 15 and 20 minutes were required to segment each slice. These high-quality annotations provided the ground truth for model validation.

In parallel, a larger set of approximately 800 weakly annotated slices was generated by untrained individuals, including graduates and undergraduates. Annotators received a brief explanation of glomerular morphology and were instructed to spend no more than 2 minutes per image. Weak annotations were generated by placing circular dot markers, with diameters of 8 pixels (about $80~\mu m$), on regions identified as glomeruli. This strategy required minimal effort per image and allowed the creation of a larger annotated dataset.

\begin{figure}[htbp]
\centering
\includegraphics[width=0.32\textwidth]{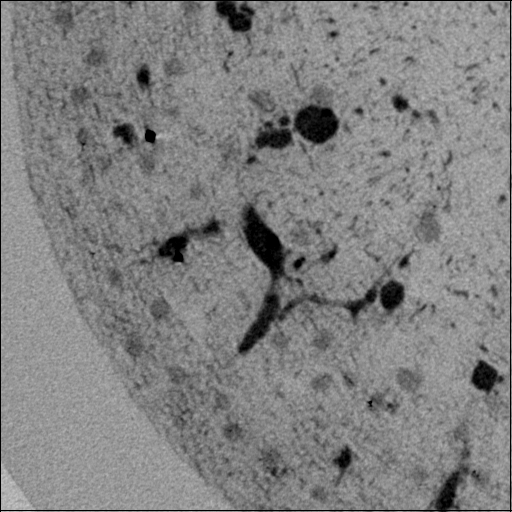}
\hfill
\includegraphics[width=0.32\textwidth]{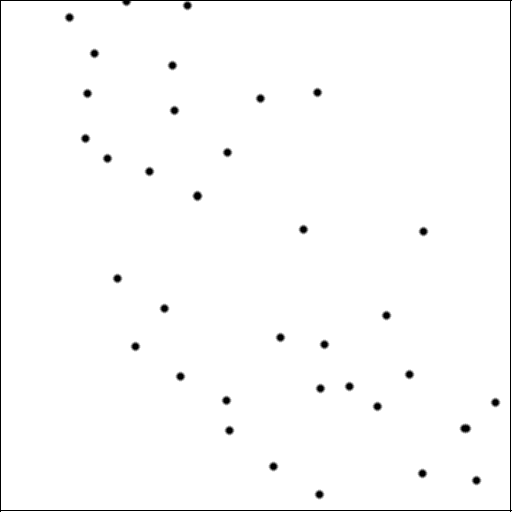}
\hfill
\includegraphics[width=0.32\textwidth]{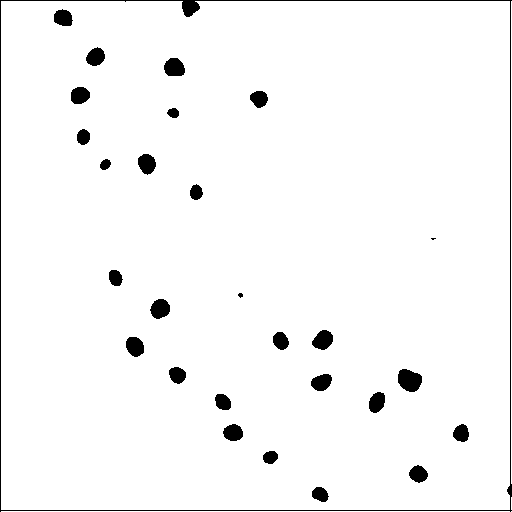}
\caption{Detail view of glomerular structures in a microCT slice (left), weak segmentation obtained from dot-based annotations by non-expert users (center), and high-quality segmentation generated by an expert-assisted procedure (right).}
\label{fig:input_prediction}
\end{figure}

The combination of a small, fully annotated dataset and a larger, weakly annotated dataset enables the training of the 2D DL model under a weakly supervised regime, reducing the annotation burden while leveraging structural priors from the high-quality subset.


\subsection{Network architecture}
\label{sec:network_architecture}

The segmentation model was implemented using the nnU-Net framework~\cite{f}. As the analysis was performed on individual microCT slices, a 2D configuration was adopted.

The network follows a U-Net~\cite{g} architecture with an encoder–decoder structure and skip connections. Convolutional blocks with batch normalization and Leaky ReLU activations are used throughout the network. The nnU-Net pipeline was configured to support the weak supervision strategy described in Section~\ref{sec:dataset_annotations}.


\subsection{Weakly supervised training strategy}
\label{sec:weak_training}

The DL model was trained using a combination of fully annotated and weakly annotated slices following a weakly supervised learning strategy. Let $\mathcal{D}_f$ denote the set of fully segmented images with pixel-wise ground-truth labels, and $\mathcal{D}_w$ denote the set of weakly annotated images, in which glomeruli locations are indicated by sparse dot annotations.

For fully annotated images, the standard Dice loss~\cite{h} was applied:
\begin{equation}
\mathcal{L}_{\text{full}} = 1 - \frac{2 \sum_i p_i g_i}{\sum_i p_i + \sum_i g_i + \epsilon},
\end{equation}
where $p_i$ is the predicted probability at pixel $i$, $g_i$ is the ground-truth label, and $\epsilon$ is a small constant (typically $10^{-7}$) introduced to avoid division by zero~\cite{i}.

For weakly annotated images, a sparse supervision loss is defined by restricting the evaluation to pixels in the vicinity of the annotated dots. Let $\Omega_w$ denote the set of pixels associated with each dot annotation. The weak Dice loss is then formulated as:
\begin{equation}
\mathcal{L}_{\text{weak}}
=
1 - \frac{2 \sum_{i \in \Omega_w} p_i}
{\sum_{i \in \Omega_w} p_i + \sum_{i \in \Omega_w} 1 + \epsilon}.
\end{equation}
which can be interpreted as encouraging high predicted probabilities in annotated regions, effectively acting as a localization constraint under sparse supervision rather than a full shape matching criterion. Unannotated regions remain unconstrained during this loss computation.

The total training loss combines the contributions from fully an weakly annotated data:
\begin{equation}
\mathcal{L} = \lambda \mathcal{L}_{\text{full}}(\mathcal{D}_f) + (1-\lambda) \mathcal{L}_{\text{weak}}(\mathcal{D}_w),
\end{equation}
where $\lambda \in [0,1]$ controls the relative contribution of strong and weak supervision. The value $\lambda = 0.5$ was chosen empirically to balance stable convergence from fully annotated samples with the exploratory learning signal provided by weak annotations. No systematic hyperparameter sweep was performed due to computational constraints, but this value consistently provided stable training behavior across experiments.

Training was performed using the Adam optimizer~\cite{j} with an initial learning rate of $1\times10^{-2}$, which was linearly decayed at each epoch. A batch size of 4 was used.

\subsection{Implementation details}
\label{sec:implementation}

All models were implemented in Python using a framework based on PyTorch. To maintain a consistent number of training samples across different training regimes, data augmentation was applied to the dataset with high-quality annotations only. Data augmentation consisted of random horizontal and vertical flips, as well as the addition of noise to the input images.

Training and inference were performed on the Leonardo high-performance computing cluster at CINECA. Jobs were executed using a single node with one task, allocated four CPU cores. The model was trained for a total training time of exactly 4 hours (approximately 400 epochs).

Inference was performed on the same computational setup. The average inference time was approximately 3 seconds per slice.


\section{Results}
\label{sec:results}

\begin{table}[b]
\centering
\caption{Results as pseudo Dice similarity coefficient (mean $\pm$ standard deviation).}
\label{tab:results_dice}
\begin{tabular}{lccc}
\hline
\textbf{Training strategy} & \textbf{Weak labels} & \textbf{High-quality labels} & \textbf{Dice score} \\
\hline
Weak annotations & 1152 (100\%) & 0 (0\%) & \DLweak{} \\
Mixed-quality annotations & 960 (~85\%) & 192 (~15\%) & \DLmix{} \\
High-quality annotations  & 0 (0\%) & 1152 (100\%) & \DLgood{} \\
\hline
\end{tabular}
\end{table}

Segmentation performance was quantitatively assessed using a pseudo Dice similarity coefficient. Three training strategies were evaluated: weak supervision using dot annotations only, mixed supervision combining weak and high-quality annotations, and fully supervised training.

Training with weak annotations alone yielded a Dice score of \DLweak{}, reflecting limited overlap with the reference segmentations. This outcome is partly due to the mismatch between the Dice loss and the dot-based labels: each dot covers only a fraction of the glomerular area, whereas the Dice metric evaluates pixel-wise overlap over the entire structure.

Mixed supervision substantially improved performance. Specifically, the model trained with 85\% weak annotations and 15\% fully segmented images achieved a Dice score of \DLmix{}, recovering a large portion of the gap toward fully supervised training. The highest accuracy was obtained by the fully supervised model, with a Dice score of \DLgood{}, representing an upper performance bound for this task. The quantitative results are reported in table~\ref{tab:results_dice}.

These results demonstrate a clear dependence of segmentation accuracy on the availability of high-quality annotations. At the same time, they indicate that a limited set of fully annotated images can significantly boost performance, providing an effective trade-off between annotation effort and segmentation quality.

\begin{figure}[h]
\centering
\includegraphics[width=0.3\textwidth]{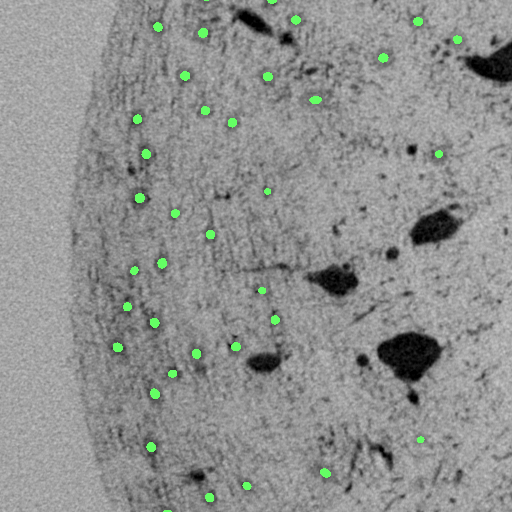}
\includegraphics[width=0.3\textwidth]{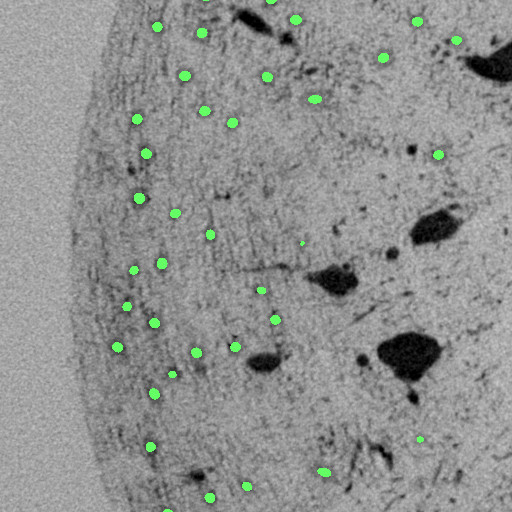}
\includegraphics[width=0.3\textwidth]{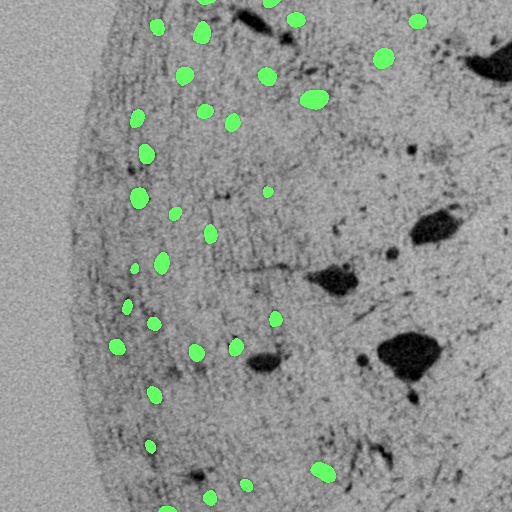}
\caption{Example of renal glomeruli segmentation predicted by models trained using only weak annotations (left), a combination of weak and fully annotated datasets (center), and only fully annotated datasets (right).}
\label{fig:overlap}
\end{figure}

\section{Conclusions}
\label{sec:conclusions}

A weakly supervised DL approach was investigated for segmenting renal glomeruli in microCT images of rat kidneys, a task characterized by small, low-contrast structures and high annotation cost. A 2D nnU-Net model was evaluated under three training regimes: weak supervision with dot-based annotations, mixed supervision combining weak and fully annotated images, and fully supervised training.

Weak-only supervision enabled the network to identify regions where glomeruli are likely to occur, even without explicit localization constraints. However, the limited accuracy observed with dot-based annotations indicates that the Dice loss may be suboptimal for this supervision type, as it evaluates pixel-wise overlap across the full object while the labels represent only sparse points. Alternative loss functions better aligned with point annotations, such as centroid- or distance-based losses, could further enhance localization and segmentation performance.

Incorporating a small number of high-quality annotations substantially improved accuracy, approaching that of the fully supervised baseline. These findings demonstrate that weak supervision provides a meaningful learning signal for anatomically relevant regions while reducing annotation effort, and highlight the potential of tailored loss formulations to improve performance in sparse annotation settings.

\acknowledgments
This work was funded by the Italian National Plan for NRRP Complementary Investments (PNC, established with the decree-law 6 May 2021, n. 59, converted by law n. 101 of 2021) in the call for the funding of research initiatives for technologies and innovative trajectories in the health and care sectors (Directorial Decree n. 931 of 06-06-2022) - project n. PNC0000003 - AdvaNced Technologies for Human-centrEd Medicine (project acronym: ANTHEM). This work reflects only the authors’ views and opinions, neither the Italian Ministry for University and Research nor the European Commission can be considered responsible for them.

The authors would like to thank Domenico Cerullo (Molecular Medicine Department) and Fabio Sangalli (Bioengineering Department) from the Istituto di Ricerche Farmacologiche Mario Negri IRCCS (Bergamo, Italy) for providing the biological samples used in this study.


\bibliographystyle{JHEP}
\bibliography{biblio.bib}


\end{document}